\documentclass{article}
\usepackage{ijcai24}

\usepackage{times}
\usepackage{soul}
\usepackage{url}
\usepackage[hidelinks]{hyperref}
\usepackage[utf8]{inputenc}
\usepackage[small]{caption}
\usepackage{graphicx}
\usepackage{amsmath}
\usepackage{amsthm}
\usepackage{booktabs}
\usepackage{algorithm}
\usepackage[switch]{lineno}

\usepackage{microtype}
\usepackage{graphicx}
\usepackage{amsmath,amssymb,amsthm}  
\usepackage{algpseudocode}
\usepackage{algorithm}  
\usepackage[algo2e]{algorithm2e}
\usepackage{academicons}
\usepackage{enumitem} 
\usepackage{amsmath, nccmath}
\usepackage{amssymb}
\usepackage{mathtools}
\usepackage{amsthm} 
\usepackage{csquotes}
\usepackage{multirow}
\usepackage[dvipsnames]{xcolor} 
\usepackage{adjustbox}
\usepackage{lipsum}
\usepackage{subfigure}
\usepackage{dblfloatfix}

\title{APALU: A Trainable, Adaptive Activation Function for Deep Learning Networks}

\author{
Barathi Subramanian$^1$ 
Rathinaraja Jeyaraj$^{1}$ 
Akhrorjon Rakhmonov Akhmadjon Ugli$^2$ \\
\affiliations
$^1$Stanford University, Palo Alto, CA-94305\\
$^2$Kyungpook National University, Daegu-41566\\ 
\emails
\{barathi1, rajaj\}@stanford.edu, akhrorrakhmonov@gmail.com 
} 

\begin{document}

\maketitle

\begin{abstract}
Activation function is a pivotal component of deep learning, facilitating the extraction of intricate data patterns. While classical activation functions like ReLU and its variants are extensively utilized, their static nature and simplicity, despite being advantageous, often limit their effectiveness in specialized tasks. The trainable activation functions also struggle sometimes to adapt to the unique characteristics of the data. Addressing these limitations, we introduce a novel trainable activation function, adaptive piecewise approximated activation linear unit (APALU), to enhance the learning performance of deep learning across a broad range of tasks. It presents a unique set of features enabling it to maintain stability and efficiency in the learning process while adapting to complex data representations. Experiments reveal significant improvements over widely used activation functions for different tasks. In image classification, APALU increases MobileNet and GoogleNet accuracy by 0.37\% and 0.04\%, respectively, on the CIFAR10 dataset. In anomaly detection, it improves the average area under the curve of One-CLASS Deep SVDD by 0.8\% on MNIST dataset, 1.81\% and 1.11\% improvements with DifferNet, and knowledge distillation, respectively, on MVTech dataset. Notably, APALU achieves 100\% accuracy on a sign language recognition task with a limited dataset. For regression tasks, APALU enhances the performance of deep neural networks and recurrent neural networks on different datasets. These improvements highlight the robustness and adaptability of APALU across diverse deep learning applications.
\end{abstract}

\section{Introduction}
The success of deep learning relies on the choice of an activation function, which imparts non-linearity to the system, allowing the network to model complex data patterns. An ideal activation function should have certain traits: a) providing non-linear curvature to fit complex patterns; b) minimizing computational complexity; and c) allowing smooth gradient flow during training, thereby enhancing the effectiveness of network training. Till today, numerous activation functions, such as rectified linear unit (ReLU) \cite{1}, leaky ReLU (LReLU) \cite{2}, exponential linear unit (ELU) \cite{3}, and gaussian error linear units (GELU) \cite{4} have been investigated for deep learning applications. Intriguingly, these activation functions have been integrated into state-of-the-art methodologies serving diverse purposes. For instance, ReLU, Leaky ReLU, and ELU have seen wide adoption in image classification \cite{5}, \cite{6}, anomaly detection \cite{7}, \cite{8}, and sign language recognition tasks \cite{9}. Similarly, GELU  have been instrumental in transformers \cite{10}, predominantly utilized for natural language processing tasks, marking significant improvements in performance. \par 

While these activation functions have demonstrated high efficacy, Ramachandran et al. \cite{11} argue that no activation function can be deemed universally optimal, underscoring the need for task-specific or adaptive activation functions. Echoing this sentiment, Agostinelli et al. \cite{12} showcased the potential for enhancing deep learning models using activation functions with trainable parameters. However, the static nature of classical activation functions, despite being advantageous, often limits their effectiveness in specialized tasks \cite{13}, \cite{14}. The trainable activation functions also struggle in applications \cite{15}, \cite{16} to adapt to the unique characteristics of the data. In response to these considerations, we introduce the adaptive piecewise approximated activation linear unit (APALU), a novel trainable activation function, designed to address the limitations of existing activation functions and offer improved performance across a broad range of deep learning tasks. It combines the strengths of linear and non-linear activation functions and includes adaptive (trainable) parameters, allowing it to dynamically learn optimal behaviour based on the given data and task. The proposed activation function is non-linear, differentiable, continuous, and monotonic, which are the properties of typical activation functions. APALU is evaluated with various benchmark datasets of different deep learning applications and compared the results with other activation functions. The results demonstrate that APALU enhances deep learning performance.  

\section{Related Works}
Numerous activation functions have been proposed over the years, starting with Sigmoid and Hyperbolic Tangent (TanH), which are the highly used activation functions in neural networks \cite{17}. The objective is to reproduce the firing behaviour of biological neurons. However, their susceptibility to the vanishing gradient problem led to the development of ReLU by Nair and Hinton \cite{1} and Maas et al. \cite{2}. As a result of its simplicity, computational efficiency, and ability to mitigate the vanishing gradient problem, ReLU gained popularity quickly. However, ReLU has some drawbacks, including the dying neuron problem. When neurons become inactive, they no longer contribute to the learning process. To address this issue, a variety of variants of ReLU have been developed. These variants include LReLU \cite{2} and Parametric ReLU \cite{18}, which allow small negative activations in the model. Clevert et al. \cite{3} have made an additional effort to adjust the dynamics of the learning process by bringing the mean activations closer to zero to enhance the learning dynamics. A search for an effective activation function  led Hendrycks and Gimpel \cite{4} to propose the GELU in the pursuit of an effective activation function. Bringing ReLUs and ELUs together ensuring fast convergence, they attempted to avoid dying ReLUs. Despite these efforts, designing an activation function that is universally effective across diverse tasks remains an open challenge. This has motivated further research into trainable and adaptive activation functions such as adaptive piecewise linear (APL) unit \cite{19}, PELU \cite{20}, TELU \cite{21} to control the slopes, biases of its piecewise linear form and to control the curvature of the negative region. However, issues like poor scalability, lack of smoothness, and optimization instability still remain. We introduce APALU to address these limitations. APALU combines adaptability and smoothness for more effective learning. 

\section{Adaptive Piecewise Approximated Activation Linear Unit (APALU)}
The APALU function is denoted by 
\begin{equation}
\label{eq:1} 
\begin{medsize}
f(x)=\left\{\begin{matrix}
a\left ( x+x\left ( \frac{1}{1+exp(-1.702(x))} \right ) \right ), & if \quad x\geqslant0 \\ 
b\left ( exp(x)-1 \right ), \quad \quad \quad \quad \quad \quad & if \quad x < 0,
\end{matrix}\right.
\end{medsize}
\end{equation} 
where $a$ and $b$ are positive trainable parameters, and $x$ is an input to the function. The initial values for parameters $a$ and $b$ are selected in random between 0 and 2. The range of an activation function is the set of output values that it can produce, given all possible input values. To determine the range of the APALU function, we examine its behaviour over the positive and negative domains separately: $x\geqslant 0 $ and $x< 0$.\\ [0.1cm]
(i) For $x\geqslant0$, given that $a > 0$ and $x\in\mathbb{R}^{+}$, the output of APALU is always non-negative. Thus, the range of $f(x)$ for $x\geqslant0$ is $0\leqslant f(x)< +\infty $, as shown in Figure 1.  \\ [0.1cm]
(ii) For $x < 0$, given that $b > 0$, the output is always less than 0 because $exp(x)$ produces non-negative outputs. Therefore, the range of $f(x)$ for $x<0$ is $-\infty < f(x) < 0  $. \par

By combining these two ranges, the range of the APALU is $f(x)\in \left ( -\infty,+\infty \right ) $, or in other words, $f(x)\in \mathbb{R}$. This broad range makes it a suitable activation function for tasks that involve predicting real-valued outputs. In the following proofs, APALU's capability of representation and convergence is justified.    

\subsection{Representation capability of APALU}
With trainable parameters $a$ and $b$, Eq. (1) can approximate any continuous function $g:[c,d]\rightarrow \mathbb{R}$ to fit higher level patterns that exist in the dataset. Considering $g$ and an arbitrary $\varepsilon > 0$, we show that there exists an APALU-based neural network ($N$), such that $\forall x \in [c,d], \left | N(x)-g(x) \right | < \varepsilon $. \\[0.15cm]
\begin{figure}[!t]
	\centering
	\includegraphics[width=0.47\textwidth]{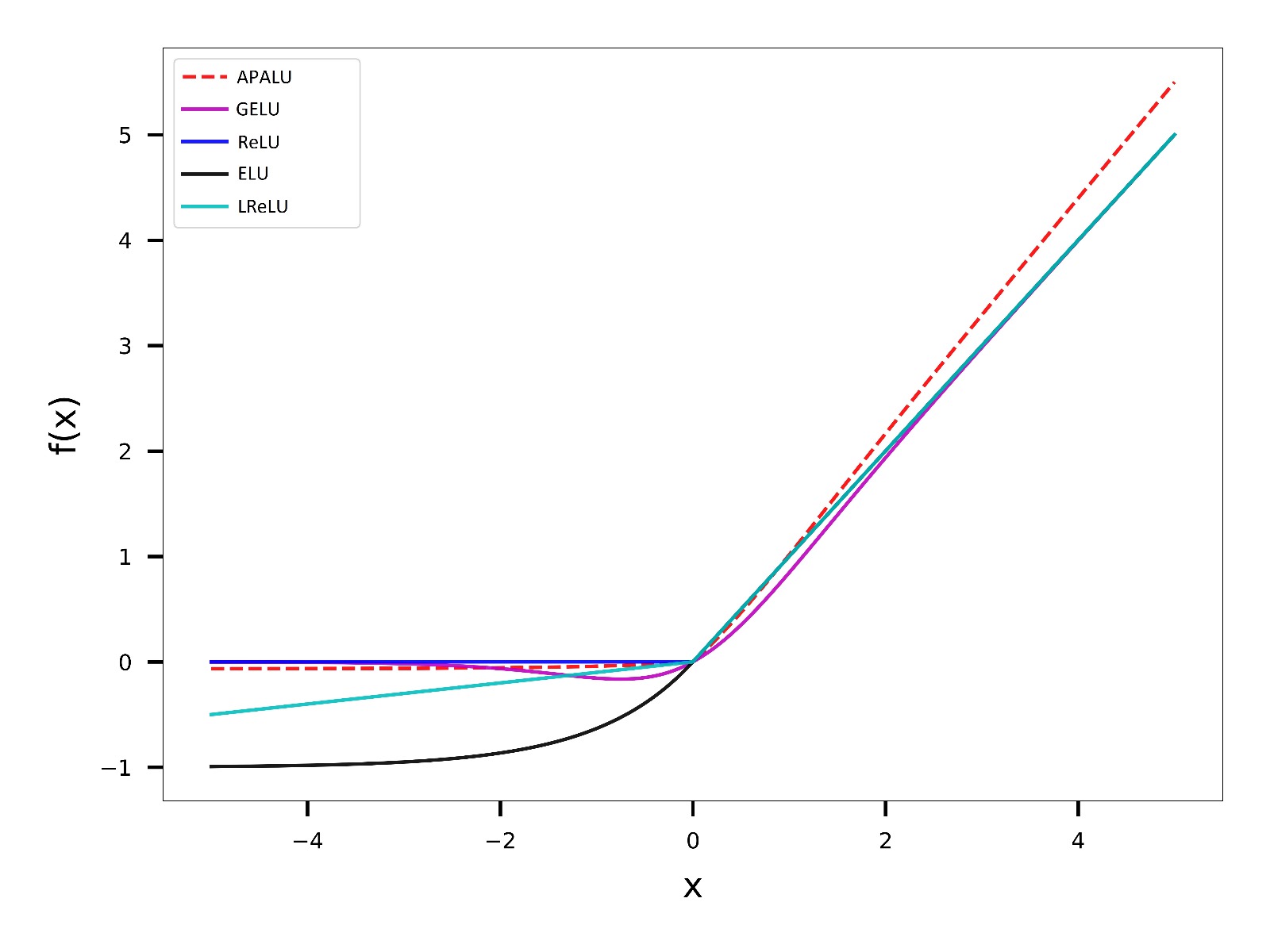}
	\caption{ APALU ($a$=0.55 and $b$=0.065), GELU, ReLU, ELU ($\alpha$ = 1), and LReLU (0.1)}
	\label{fig:1}
\end{figure} 
\textit{Proof 1:} The function $f(x)$ is piecewise, with $f_{pos}(x)=a\left ( x+x\left ( \frac{1}{1+exp(-1.702(x))} \right ) \right )$ modelling behavior for $x\geq 0$ and $f_{neg}(x)=b(exp(x)-1)$ for $x<0$. The $f_{pos}(x)$ term exhibits a sigmoidal characteristic due to $\frac{1}{1+exp(-1.702(x))}$, allowing for modeling of S-shaped curves in $g(x)$ for positive $x$. Similarly, $f_{neg}(x)$ shows an exponential growth pattern, suitable for approximating segments of $g(x)$ that display rapig growth or deacy for negative $x$. \par

Assume $M$ number of APALU unit in $N$ and each unit is an instaniation of $f(x)$ with particular $a$ and $b$ parameters. By the universal approximation theorem, such a network can approximate any continuous function on a compact interval within any bound of error $\varepsilon $, given a sufficiently large $M$. For each APALU unit in $N$, select $a$ and $b$, such that the combined output of these units closely follows the shape of $g(x)$ on the interval $[c,d]$. By fine-tuning these parameters, align the piecewise segments of each APALU unit to replicate the behavior of $g(x)$ across its domain. With appropriate selection of M and tuning of $a$ and $b$, it ensures that the inequality $\left | N(x)-g(x) \right | < \varepsilon $ holds $\forall x \in [c,d]$. \par

Therefore, it is proved that the APALU function can approximate any continuous function on a closed interval to any desired degree of accuracy, demonstrating its robust representation capability. This conclusion validates the potential of APALU in versatile function approximation within neural network architectures.

\subsection{Convergence rate in learning}
In a neural network employing APALU function, the convergence rate during the learning process is faster compared to networks utilizing traditional activation functions like ReLU or sigmoid, under specific conditions.  \\[0.15cm]
\textit{Proof 2:} Consider a neural network trained to minimize a loss function $J(w)$ using gradient descent, where $w$ represents the learnable parameters of the network. The gradient descent update rule is:
\begin{equation}
\label{eq:2} 
w^{(t+1)}=w^{(t)}-\eta \nabla J(w^{(t)}),
\end{equation}
where $\eta$ is the learning rate. By applying the mean value theorem, we have

\begin{equation}
\left | \nabla J(w^{(t)}) \right | \leqslant M \left | \frac{\partial J}{\partial w} \right |,
\end{equation}
where $M$ is a constant dependent on the bounds of the derivative of the activation function used in the network. For a network using APALU function, the derivative of the loss function with respect to the weights $\left | \frac{\partial J}{\partial w} \right |$ is upper bounded by a constant $K$ due to the characteristics of the APALU. This results in the gradient being bounded as
\begin{equation}
\left | \nabla J(APALU(w^{(t)})) \right | \leqslant K.
\end{equation}
In contrast, for networks employing traditional activation functions like ReLU or sigmoid, the derivative of the loss function with respect to the weights can grow significantly larger, potentially leading to greater gradient magnitudes. \par

As a result, the gradient magnitude $\left | \nabla J(w^{(t)}) \right |$ tends to remain lower for networks with APALU function across training iterations compared to networks using traditional activation functions. The properties of gradient descent suggest that a smaller gradient magnitude leads to accurate corvergence in the loss function $J(w^{(t)})$, when the gradient is close to the bottom of the function. Therefore, the neural networks employing APALU function demonstrate a faster convergence rate compared to networks using traditional activation functions like ReLU or sigmoid, which might cause jumps around lower point of the loss function.

\subsection{Vanishing gradient robustness}
Neural networks utilizing APALU function exhibit greater robustness against vanishing gradients compared to networks with traditional activation functions like sigmoid or ReLU, especially in deep neural network architectures.  \\[0.15cm]
\textit{Proof 3:} In deep neural networks, the vanishing gradient problem is characterized by the exponential decrease in the gradient magnitude $\left ( \left | \frac{\partial J}{\partial w} \right |   \right )$ as it is propagated back through layers during training, leading to ineffective learning in earlier layers. For APALU function, the derivative of $f(x)$ with respect to its input $x$ is lower bounded by a positive constant $\alpha = a e^{-1.702u}$, where $a$ and $u$ are parameters of APALU. This lower bound ensures that the gradient does not diminish to negligible values through layers. Traditional activation functions like sigmoid suffer from saturation, where the derivative of $J$ approaches zero for large magnitude inputs, and ReLU has a derivative of zero for all negative inputs. These properties can lead to extremely small gradients in the deeper layers of a network. \par
In a deep network with L layers using APALU, the chain rule for backpropagation implies that the gradient of the loss with respect to the weights in the first layer involves the product of L derivatives. Due to the lower bound on the derivatives in APALU networks, this product maintains a sufficient magnitude, preventing the gradient from diminishing exponentially, a common issue in networks with sigmoid or ReLU activations.

\subsection{Approximation capability}
Given a continuous function $g:[c,d]\rightarrow \mathbb{R}$ and an accuracy level $\varepsilon > 0$, there exist an APALU network $f(x;\theta)$ parameterized by $\theta$, such that: $sup_{x\in [a,b]}\left | f(x;\theta) -g(x)\right |<\varepsilon$. \\[0.15cm]
\textit{Proof 4:} By applying the Stone-Weierstrass theorem, for any $g(x)$ defined on a closed interval $[a,b]$ and for accuracy $\varepsilon>0$, there is a polynomial function $p(x)$ that uniformly approximates $g(x)$ within $\varepsilon/2$ over $[a,b]$. This theorem is fundamental in approximation theory and provides a basis for approximating continuous functions using polynomials. \par

According to Proof 1, the APALU function can represent any piecewise nonlinear shape with a high level of precision. This ability of APALU to closely mimic complex patterns is crucial for the next step of the proof. Construct an APALU network $f(x;\theta)$ with parameters $\theta$ that include a specified number of pieces $K$ and corresponding $\{a; b\}$. These parameters are chosen such that the APALU network approximates the $p(x)$ within an error margin of $\varepsilon/2$. The choice of $K$, $a$ and $b$ is critical for ensuring that the APALU network can closely follow the shape of the polynomial approximation. \par

By using the triangle inequality, the overall approximation error between the APALU network $f(x;\theta)$ and the original function $g(x)$ can be bounded. Specifically, the inequality $\left | f(x;\theta)-g(x) \right |\leqslant \left | f(x;\theta)-p(x) \right |+\left | p(x)-g(x) \right |$ holds for all $x$ in $[a,b]$. Since each of the terms on the right-hand side is less than $\varepsilon/2$, their sum is less than $\varepsilon$. Therefore, it follows that $sup_{x\in [a,b]}\left | f(x;\theta) -g(x)\right |<\varepsilon$. This proof demonstrates that an appropriately parameterized APALU network $f(x;\theta)$ can approximate any given continuous function $g(x)$ over the interval $[a,b]$ within any desired degree of accuracy $\varepsilon$. This establishes the powerful approximation capabilities of the APALU network, making it a versatile tool in neural networks.

\section{Experiments}
We comprehensively analyse the effectiveness of APALU compared to four standard activation functions (ReLU, LReLU, ELU, and GELU) on different tasks, such as image classification, anomaly detection, sign language recognition, and regression tasks. The experimental results show that APALU excels in most network settings compared to the standard activations. For all our experiments, values for a and b in APALU are initialized and then refined via the backpropagation \cite{22}.

% \paragraph{Selection of $a$ and $b$:} The value for parameters $a$ and $b$ is selected random between 0 and 2. This provides a reasonable starting point for the parameters. These hyperparameters are adjusted to identify suitable values for different models. The procedure is to fix $a$ and increase/decrease $b$ with 0.01 or fix $b$ and increase/decrease $a$ with 0.01. Based on the validation accuracy on 5 to 10 epochs, the right combination is chosen by further increasing/decreasing a and b with 0.01 on a few trials for stable convergence. 

\subsection{Image Classification}
We evaluate the APALU for image classification on the MNIST \cite{23} and CIFAR-10 \cite{24} datasets. Four prominent CNN architectures are employed: MobileNet \cite{25}, GoogleNet \cite{26}, SENet18 \cite{27}, and ResNet50 \cite{6}.  

\paragraph{MNIST:} We first evaluate APALU on the MNIST dataset that contains 60k training and 10k testing grey-scale images of size 28x28 with ten different classes. The batch size and learning rate are set to 128 and 0.001, respectively. Data augmentation techniques, such as random cropping and horizontal flipping are applied along with normalization in the data preprocessing stage. Adam optimizer and cross-entropy loss are used for the training process. The APALU is used in all the layers (except Softmax) and trained for 100 epochs. Table 1 demonstrates the superiority of our proposed function, outpacing all baseline activation functions across different network models. This advantage is not only reflected in higher performance but also in the stability of the results, as indicated by the mean $\pm $ standard deviation values.

\begin{table} [H]  
\scriptsize
\centering
\caption{Test accuracy (in \%) of the models with different activation functions on MNIST.}
\label{table2}
\begin{tabular}{|l|c|c|c|c|}
\hline
\multicolumn{1}{|c|}{\textbf{\begin{tabular}[c]{@{}c@{}}Activation \\ Function\end{tabular}}} & \textbf{MobileNet}                                                                  & \textbf{GoogleNet}                                                               & \textbf{ResNet50}                                                                & \textbf{SENet18}                                                                 \\ \hline
\textbf{ReLU}                                                                                 & 98.95±0.33                                                                          & 99.59±0.13                                                                       & 99.47±0.11                                                                       & 99.52±0.11                                                                       \\ \hline
\textbf{LReLU}                                                                                & 99.01±0.26                                                                          & 99.44±0.16                                                                       & 99.47±0.12                                                                       & 99.53±0.08                                                                       \\ \hline
\textbf{ELU}                                                                                  & 99.29±0.17                                                                          & 99.53±0.11                                                                       & 99.42±0.09                                                                       & 99.47±0.06                                                                       \\ \hline
\textbf{GELU}                                                                                 & 99.14±0.20                                                                          & 99.54±0.13                                                                       & 99.49±0.04                                                                       & 99.54±0.06                                                                       \\ \hline
\textbf{APALU}                                                                       & \begin{tabular}[c]{@{}c@{}}\textbf{99.32±0.16}\\ (a=1.05, \\b=1.20)\end{tabular} & \begin{tabular}[c]{@{}c@{}}\textbf{99.63±0.12}\\ (a=1.15, \\b=1.30)\end{tabular} & \begin{tabular}[c]{@{}c@{}}\textbf{99.51±0.02}\\ (a=0.25, \\b=0.93)\end{tabular} & \begin{tabular}[c]{@{}c@{}}\textbf{99.57±0.04}\\ (a=0.35, \\b=0.93)\end{tabular} \\ \hline
\end{tabular}
\end{table}  

\paragraph{CIFAR10:} We extend our evaluation to the CIFAR-10 dataset, a more complex collection of 50,000 training and 10,000 testing images of size 32×32 RGB with ten classes. In this context, the experimental setup is the same as used for MNIST. The results, as given in Table 2, are obtained from the mean of five different runs. Notably, the APALU maintains its edge, consistently achieving the highest accuracies. Its performance across various models reaffirms APALU's robustness, adaptability, and effectiveness in handling the increased complexity and variety in the CIFAR-10 dataset, further consolidating its superiority over traditional classical functions.

\begin{table} [H]  
\scriptsize
\centering
\caption{Test accuracy (in \%) of the models with different activation functions on CIFAR10.}
\label{table3}
\begin{tabular}{|l|c|c|c|c|}
\hline
\multicolumn{1}{|c|}{\textbf{\begin{tabular}[c]{@{}c@{}}Activation   \\ Function\end{tabular}}} & \textbf{MobileNet }                                                                 & \textbf{GoogleNet }                                                               & \textbf{ResNet50 }                                                               & \textbf{SENet18 }                                                                  \\ \hline
\textbf{ReLU}                                                                                   & 90.10±0.22                                                                         & 93.43±0.48                                                                       & 93.74±0.34                                                                      & 93.70±0.16                                                                        \\ \hline
\textbf{LReLU}                                                                                  & 90.10±0.19                                                                         & 89.28±0.82                                                                       & 93.83±0.42                                                                      & 93.66±0.19                                                                        \\ \hline
\textbf{ELU}                                                                                    & 90.92±0.25                                                                         & 92.47±0.76                                                                       & 93.53±0.66                                                                      & 93.39±0.20                                                                        \\ \hline
\textbf{GELU}                                                                                   & 90.71±0.20                                                                         & 93.16±0.61                                                                       & 93.81±0.46                                                                      & 93.72±0.18                                                                        \\ \hline
\textbf{APALU}                                                                           & \begin{tabular}[c]{@{}c@{}}\textbf{91.09±0.06}\\ (a=1.05, \\ b=1.20)\end{tabular} & \begin{tabular}[c]{@{}c@{}}\textbf{93.73±0.12}\\ (a=1.15, \\ b=1.30)\end{tabular} & \begin{tabular}[c]{@{}c@{}}\textbf{93.89±0.25}\\ (a=0.35, \\  b=0.95\end{tabular} & \begin{tabular}[c]{@{}c@{}}\textbf{93.75±0.17}\\  (a=0.15,  \\ b=0.93)\end{tabular} \\ \hline
\end{tabular}
\end{table} 
\paragraph{Parameter ($a$ and $b$) for MNIST and CIFAR10:} The careful selection of initial value for the parameters $a$ and $b$ (between 0 and 2) in APALU showcases the ability to adapt to both the simple contours of MNIST and the intricate patterns within CIFAR-10 across various architectures. For the MNIST dataset, the tuning is driven by the specific learning dynamics and characteristics of the models. MobileNet's lightweight nature was complemented by $a$=1.05 and $b$=1.20, while GoogleNet is benefited from more non-linear transformations with $a$=1.15 and $b$=1.30. The ResNet50's need for more linear behaviour led to $a$=0.25 and $b$=0.93, and SENet18's attention mechanism found balance with $a$=0.35 and $b$=0.93. Transitioning to CIFAR-10's complex patterns, MobileNet's detection of fine textures is optimized with $a$=1.05, $b$=1.20, while GoogleNet's hierarchical pattern recognition is enhanced with $a$=1.15, $b$=1.30. ResNet50's deep connections are aligned with $a$=0.35, $b$=0.95, and SENet18's channel recalibration was harmonized with $a$=0.15, $b$=0.93. The distinct $a$ and $b$ values for each architecture on both the datasets not only showcase APALU's adaptability but also its ability to self-tune according to the unique demands and complexities of each model. This customization leads to enhanced and consistent performance, demonstrating the activation function's robustness and compatibility across varying architectures.

\subsection{Anomaly Detection}
Anomaly detection is an important problem in deep learning. In this section, we evaluate our proposed activation function on real-world defect detection problem. We use the MVTec AD \cite{28} dataset designed to test anomaly localization algorithms for industrial quality control. The difficulty in these datasets lies in the similarity of anomalies and normal examples. To the best of our knowledge, MVTec AD is the only publicly available multi-object and multi-defect anomaly dataset. It contains 5354 high-resolution color images of 10 object and 5 texture categories. The number of training samples per category ranges from 60 to 320, which is challenging for the estimation of the distribution of normal samples. The anomalies differ in their size, shape and structure and thus cover several scenarios in industrial defect detection. To show the efficiency of APALU, it is compared with models, such as DifferNet \cite{29}, PaDiM \cite{30}, PatchCore \cite{8}, and multiresolution knowledge distillation (KD) \cite{31}.

\begin{figure*}[!b]
	\centering
	\includegraphics[width=0.95\textwidth]{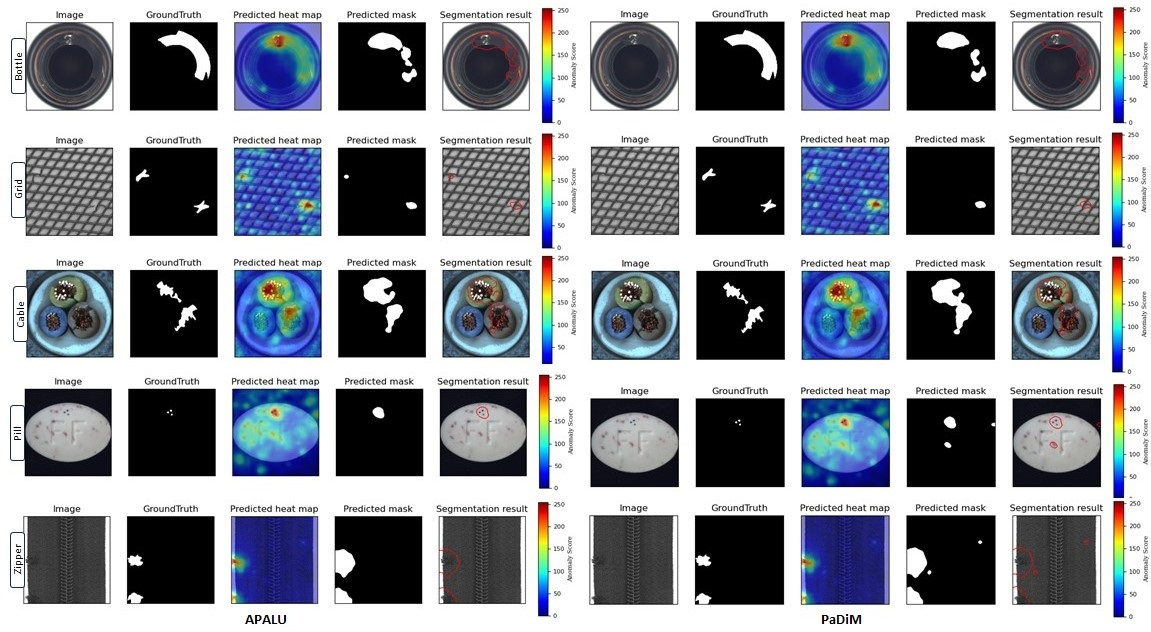}
	\caption{ Anomaly localization for PaDiM\_baseline(right) and PaDiM\_APALU (left) models on MVTec AD dataset.}
	\label{fig:2}
\end{figure*}

\paragraph{Observations on DifferNet and KD network models:} DifferNet is designed to detect subtle differences in patterns, making it suitable for anomaly detection. In the modified version, the ReLU is replaced with APALU, having initial values $a$=0.55 and $b$=0.065. Similarly, KD is an architecture that aims to transfer knowledge from a complex model to a simpler one. Like DifferNet, the KD model was also tested with APALU replacing ReLU, using the same initial values. For the experiment, we follow the same baseline configuration as mentioned in \cite{29}, \cite{31} to have fair comparison between the baseline models and with the incorporation of the APALU in both the network models. The findings are summarized in Table 3, which showcases the average image-level accuracy for different object categories, both for the baseline configurations and the adaptations using APALU. \par

The adaptation of the DifferNet model with APALU shows a consistent improvement across various classes, achieving an overall average accuracy of 95.17\%, compared to the 
\begin{table} [t!]  
\scriptsize
\centering
\caption{Comparison of area under the ROC (in \%) for anomaly detection across various categories of MVTec AD using DifferNet and KD models, with and without APALU ($a$=0.55, $b$=0.065). }
\label{table4}
\begin{tabular}{|l|c|c|c|c|}
\hline
\multicolumn{1}{|c|}{\textbf{Class}} & \textbf{\begin{tabular}[c]{@{}c@{}}DifferNet \\ (Baseline)\end{tabular}} & \textbf{\begin{tabular}[c]{@{}c@{}}DifferNet\\ (APALU)\end{tabular}} & \textbf{\begin{tabular}[c]{@{}c@{}}KD \\ (Baseline)\end{tabular}} & \textbf{\begin{tabular}[c]{@{}c@{}}KD\\ (APALU)\end{tabular}} \\ \hline
\textbf{Grid}                        & 79.9                                                                    & \textbf{81.2}                                                           & 75.5                                                             & \textbf{79.3}                                                    \\ \hline
\textbf{Leather}                     & 97.1                                                                    & \textbf{98.7}                                                           & 92.0                                                             & \textbf{94.5}                                                    \\ \hline
\textbf{Tile}                        & 99.4                                                                    & \textbf{99.9}                                                           & 89.8                                                             & \textbf{90.6}                                                    \\ \hline
\textbf{Carpet}                      & 91.7                                                                    & \textbf{92.8}                                                           & 77.7                                                             & \textbf{81.2}                                                    \\ \hline
\textbf{Wood}                        & 99.7                                                                    & \textbf{99.8}                                                           & 93.7                                                             & \textbf{94.9}                                                    \\ \hline
\textbf{Bottle}                      & 99.0                                                                    & \textbf{99.4}                                                           & 99.3                                                             & \textbf{99.4}                                                    \\ \hline
\textbf{Capsule}                     & 86.9                                                                    & \textbf{88.4}                                                           & 81.0                                                             & \textbf{81.4}                                                    \\ \hline
\textbf{Pill}                        & 88.8                                                                    & \textbf{90.9}                                                           & 83.3                                                             & \textbf{83.9}                                                    \\ \hline
\textbf{Transistor}                  & 91.1                                                                    & \textbf{91.4}                                                           & 86.0                                                             & \textbf{88.4}                                                    \\ \hline
\textbf{Zipper}                      & 95.1                                                                    & \textbf{95.3}                                                           & 94.0                                                             & \textbf{94.1}                                                    \\ \hline
\textbf{Cable}                       & 95.9                                                                    & \textbf{96.4}                                                           & 88.6                                                             & \textbf{89.8}                                                    \\ \hline
\textbf{Hazelnut}                    & 99.3                                                                    & \textbf{99.9}                                                           & 97.9                                                             & \textbf{98.0}                                                    \\ \hline
\textbf{Metal Nut}                   & 96.1                                                                    & \textbf{96.8}                                                           & 74.1                                                             & \textbf{74.1}                                                    \\ \hline
\textbf{Screw}                       & 96.3                                                                    & \textbf{96.7}                                                           & 81.4                                                             & \textbf{90.6}                                                    \\ \hline
\textbf{Toothbrush}                  & 99.4                                                                    & \textbf{100.0}                                                          & 91.9                                                             & \textbf{93.1}                                                    \\ \hline
\textbf{Average}                     & 94.38                                                                   & \textbf{95.17}                                                          & 87.08                                                            & \textbf{88.89}                                                   \\ \hline
\end{tabular}
\end{table} 
baseline performance of 94.38\%. Notable improvements are observed in classes such as "Leather" and "Tile." Similar enhancements are noted with the KD model when APALU is integrated. While KD model yields an average accuracy of 87.08\%, the APALU pushes the performance to 88.89\%. The increase is particularly significant in categories like "Screw" and "Wood." This adaptive activation function is specifically tailored with initial values $a$=0.55 and $b$=0.065 to correspond to the unique patterns and characteristics evident in the MVTec dataset, which encompasses diverse anomalies and normal examples. The choice of these values is emerged from a meticulous study of the distribution of the MVTec data, leading to more precise anomaly detection and localization. The results underscore the potential of APALU as a versatile activation function and illustrate the importance of thoughtful parameter selection in enhancing the efficacy of deep learning models for such complex tasks.

\begin{table} [t!]  
\scriptsize
\centering
\caption{Comparative analysis of average image-level and pixel-level accuracy for anomaly detection using PaDiM and PatchCore models (Baseline and APALU)}
\label{table5}
\begin{tabular}{|l|c|c|}
\hline
\multicolumn{1}{|c|}{\textbf{Model}}                                    & \textbf{\begin{tabular}[c]{@{}c@{}}Average Image \\ level accuracy\end{tabular}} & \textbf{\begin{tabular}[c]{@{}c@{}}Average pixel   \\ level accuracy\end{tabular}} \\ \hline
\begin{tabular}[c]{@{}l@{}}PaDiM \\ (Baseline)\end{tabular}              & 90.1                                                                              & 96.5                                                                                 \\ \hline
\textbf{\begin{tabular}[c]{@{}l@{}}PaDiM\\ (APALU)\end{tabular}}     & \textbf{91.0}                                                                     & \textbf{96.6}                                                                        \\ \hline
\begin{tabular}[c]{@{}l@{}}Patchcore \\ (Baseline)\end{tabular}          & 98.2                                                                              & 97.6                                                                                 \\ \hline
\textbf{\begin{tabular}[c]{@{}l@{}}Patchcore\\ (APALU)\end{tabular}} & \textbf{98.3}                                                                     & \textbf{97.7}                                                                        \\ \hline
\end{tabular}
\end{table} 

\paragraph{Observations on PaDiM and PatchCore model:} Anomaly detection models, PaDiM and PatchCore, with and without APALU, also show significant distinctions. PatchCore exhibits a remarkable performance, significantly outpacing PaDiM in both image-level and pixel-level accuracy. The average image-level accuracy for PatchCore reaches 98.2\% in the baseline configuration and it is slightly improved to 98.3\% with APALU, whereas PaDiM records 90.1\% and 91.0\%, as shown in Table 4, respectively. A similar trend is observed in pixel-level accuracy, where PatchCore achieves an average of 97.6\% in the baseline and 97.7\% with APALU, compared to 96.5\% and 96.6\% for PaDiM. These results not only underscore the robustness and effectiveness of the PatchCore model in anomaly detection but also demonstrate the subtle enhancement brought about by the integration of APALU. The combination of local patch features, coreset-reduced patch-feature memory bank, and specific adaptations in PatchCore appears to contribute to its superior performance in the industrial quality control setting. \par 

Figure 2 illustrates the anomaly localization results of the models, PaDiM\_baseline (right) and PaDiM\_APALU (left), tested on the MVTec AD dataset. Visually, the PaDiM\_APALU model demonstrates a noticeable enhancement in localization accuracy. The implementation of APALU seems to refine the detection contours, highlighting the anomalous regions with greater precision. This leads to a more coherent and clear delineation of the anomalies, allowing for better interpretation and validation against the ground truth.

\begin{table} [b!]  
\scriptsize
\centering
\caption{Performance of APALU with sign language recognition task.}
\label{table6}
\begin{tabular}{|l|c|c|c|c|c|}
\hline
\multicolumn{1}{|c|}{\textbf{Network Model}} & \textbf{MAE} & \textbf{MSE} & \textbf{$R^{2}$} & \textbf{Loss} & \textbf{Accuracy(\%)} \\ \hline
MOPGRU (Baseline)                            & 0.22         & 1.34         & 0.88             & 0.22               & 95                         \\ \hline
MOPGRU (APALU)                            & 0.0          & 0.0          & 1.0              & 0.02               & 100                        \\ \hline
\end{tabular}
\end{table} 

\subsection{Sign Language Recognition}
Sign language recognition is a critical task in human-computer interaction, enabling seamless communication with individuals who are deaf. For the experiments, the configuration is set according to \cite{9}. The authors employ mediapipe optimized gated recurrent unit (MOPGRU)  model to improve the performance with limited datasets. The performance is further enhanced with APALU. The assessment is done based on mean absolute error (MAE), mean squared error (MSE), $R^2$(coefficient of determination), test loss, and test accuracy. As shown in Table 5, the baseline MOPGRU model displays a commendable performance with MAE of 0.22, MSE of 1.34,$ R^2$ score of 0.88, test loss of 0.22, and test accuracy of 95\%. While the results are promising, there is still room for improvement, particularly in accuracy and error metrics. When APALU is used in MOPGRU, the results show a remarkable improvement across all metrics. The MAE and MSE were reduced to zero, and $R^2$ reaches to 1, indicating that the model's predictions were in complete alignment with the actual values in the dataset. Furthermore, the test loss is significantly minimized, and the test accuracy achieved a 100\%. Part of this success intricate anomalies in the MVTec dataset, where both models are tasked with discerning subtle variations in defects. The selected values allow APALU to finely tune its response, optimizing the sensitivity of both models to these nuanced the differences. The precision in choosing these values for $a$ and $b$ demonstrates a profound understanding of the underlying can be attributed to the careful selection of initial values for the APALU parameters, with $a$=1.01 and $b$=1.00. These exceptional results with APALU can be attributed to its ability to adapt to the specificities of the sign language recognition task, especially with limited datasets. By capturing complex spatial and temporal patterns within video frames, the APALU-enabled model delivered more accurate and precise predictions. \par

Figure 3 illustrates the learning graph comparison of proposed activation function with other three activation functions for the MOPGRU model applied to the sign language recognition task. The curve associated with APALU stands out prominently, showing a consistent decrease in training loss and a corresponding increase in training accuracy, indicating a more efficient and effective learning process compared to the other activation functions. The APALU's adaptability and fine-tuned parameterization seem to contribute to its superior performance. In contrast, the curves for ReLU, ELU, and LReLU present varying degrees of convergence and stability, which are less smooth and gradual unlike APALU. The differences in these curves underscore the unique benefits of APALU, especially its ability to handle the complex spatial and temporal dynamics inherent in sign language recognition. In summary, the learning graph comparison provides a clear visual validation of the superiority of APALU over classical activation functions in the context of the MOPGRU model for sign language recognition.

\begin{figure}[t!]
    \includegraphics[width=.5\textwidth]{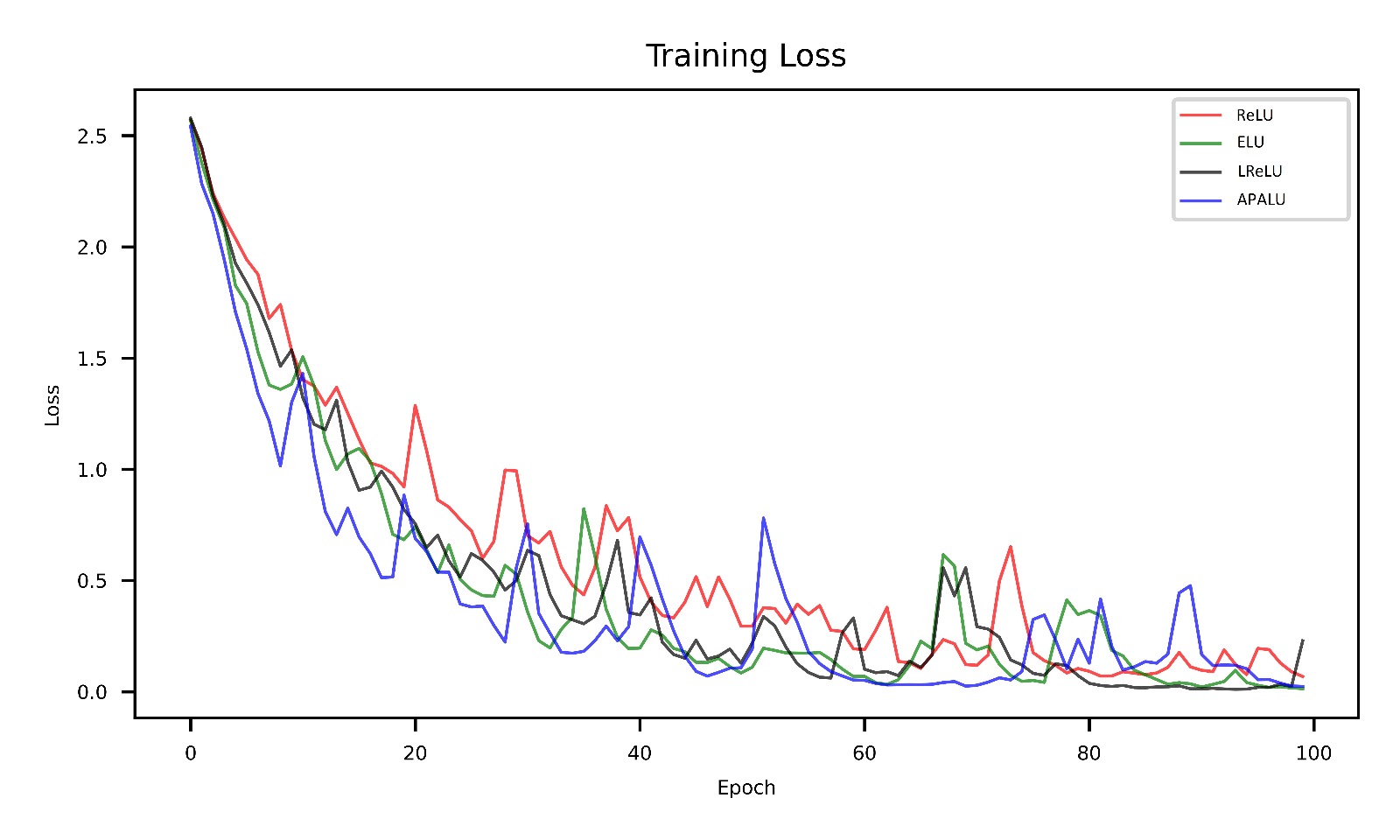}\\
    \includegraphics[width=.5\textwidth]{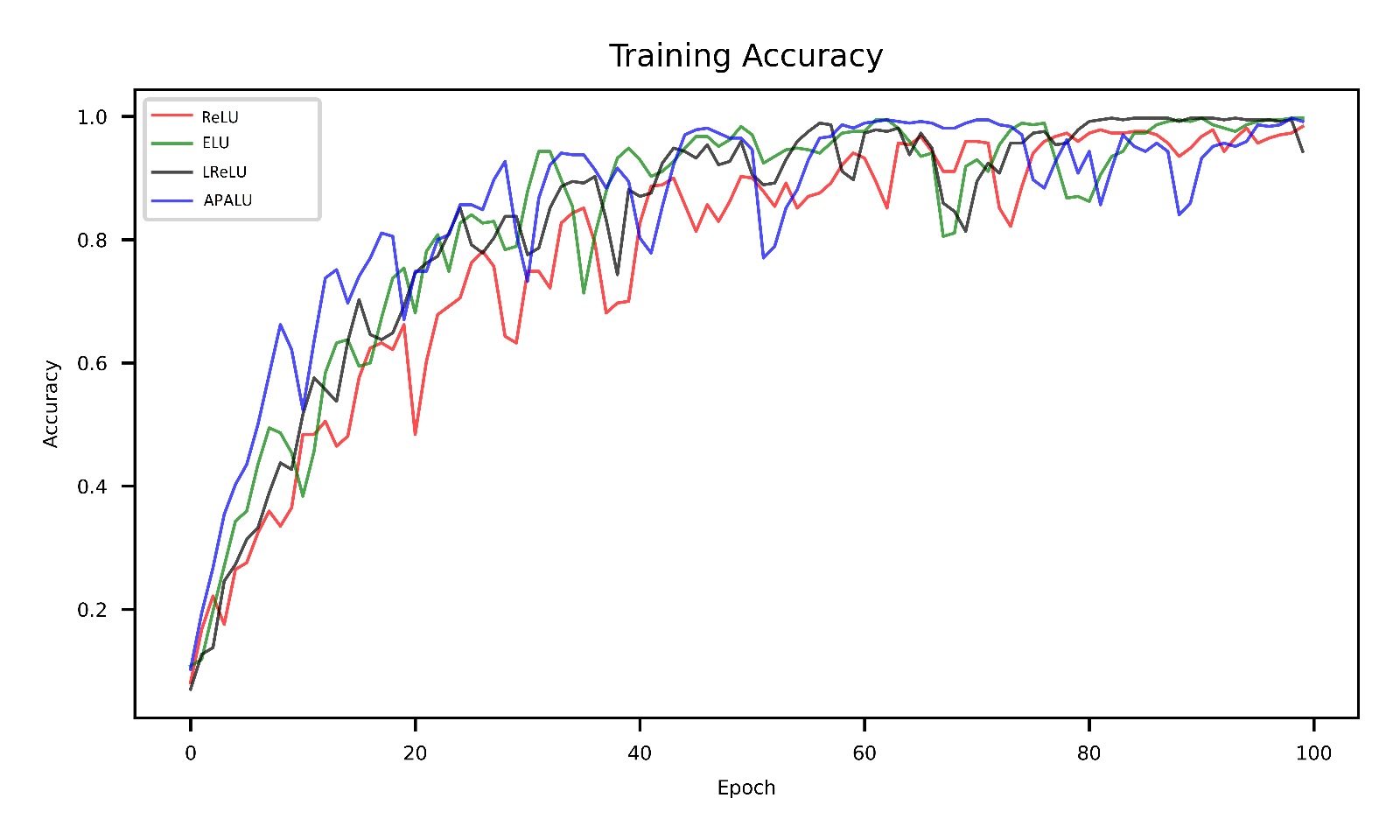}
    \caption{Training loss and accuracy with different activation functions on MOPGRU model.}
 \end{figure} 
 
% \begin{figure}[!t]
% 	\centering
% 	\includegraphics[width=0.4\textwidth]{Fig 3.jpg}
% 	\caption{Training loss and test accuracy for MOPGRU model with different activation functions for sign language recognition task.}
% 	\label{fig:3}
% \end{figure} 

\subsection{Regression and Stock Forecasting}
To show the efficacy of the APALU with numerical predictions, it is empirically evaluated with regression task using deep neural networks (DNN) on Boston house prediction \cite{32} dataset, and stock forecasting task using gated recurrent unit (GRU) on Yahoo finance stock price prediction (Yfinance) dataset retrieved in real-time using the library. Boston house prediction dataset consists of 506 instances, each with 13 numerical features, such as crime rate, average number of rooms, and accessibility to highways. This dataset is commonly utilized for developing and benchmarking regression models. Yahoo finance stock price prediction dataset includes various financial metrics like opening price, closing price, trading volume, and adjusted closing price. This dataset allows the modelling of complex time series patterns in stock prices, making it suitable for evaluating forecasting algorithms. In the regression task, the performance of three different activation functions is evaluated using two key metrics: training cost and test cost. The results, as shown in Table 6, emphasize the performance of APALU in reducing both training and testing costs, indicating that it can model the underlying relationship in the observed data more effectively. In comparison, ReLU also provides good performance but falls slightly short of APALU. On the other hand, the TanH activation function appears to be less suitable for this specific regression task, as evidenced by its higher costs. In addition, the values of $a$=0.55 and $b$=0.065 (for APALU) for the regression task plays a crucial role in this success. In addition, the batch size for training is 256 and executed for 200 epochs. \par

\begin{table} [H]  
\scriptsize
\centering
\caption{Training and test cost performance on Boston dataset.}
\label{table7}
\begin{tabular}{|l|c|c|}
\hline
\textbf{Activation Function}             & \textbf{Training cost}                 & \textbf{Test cost}                    \\ \hline
Tanh                                     & 11.044                                 & 91.81                                 \\ \hline
ReLU                                     & 13.244                                 & 24.08                                 \\ \hline
{\color[HTML]{000000} \textbf{APALU}} & {\color[HTML]{000000} \textbf{12.538}} & {\color[HTML]{000000} \textbf{23.11}} \\ \hline
\end{tabular}
\end{table} 

 From Figure 4, it is evident that the validation and training loss for APALU is significantly better when compared to other activation functions like TanH and ReLU without any overfitting issues. For Yahoo finance stock forecasting, the performance of activation functions is evaluated using mean absolute percentage error (MAPE) and root mean squared error (RMSE). As shown in Table 7, the performance of APALU is significantly better than activation functions, achieving the lowest MAPE and RMSE, highlighting its effectiveness in handling the complexity of financial time series data. This success can be attributed to the specific parameterization of $a$=0.40 and $b$=1.00 as initial trainable values for the stock forecasting task. The choice of these values allows APALU to capture underlying patterns and trends more accurately. 

 \begin{table} [H]  
\scriptsize
\centering
\caption{Mean and standard deviation of MAPE and RMSE on Yfinance dataset with different activation functions.}
\label{table8}
\begin{tabular}{|l|c|c|}
\hline
\textbf{Activation Function}             & \textbf{MAPE}                            & \textbf{RMSE}                              \\ \hline
ReLU                                     & 0.03±0.0                                 & 550.8±140.9                                \\ \hline
GELU                                     & 0.02±0.0                                 & 350.6±89.9                                 \\ \hline
{\color[HTML]{000000} ELU}               & {\color[HTML]{000000} 0.02±0.0}          & {\color[HTML]{000000} 384.5±76.4}          \\ \hline
{\color[HTML]{000000} \textbf{APALU}} & {\color[HTML]{000000} \textbf{0.01±0.0}} & {\color[HTML]{000000} \textbf{272.6±63.5}} \\ \hline
\end{tabular}
\end{table}

\begin{figure}[t!]
    \includegraphics[width=.47\textwidth]{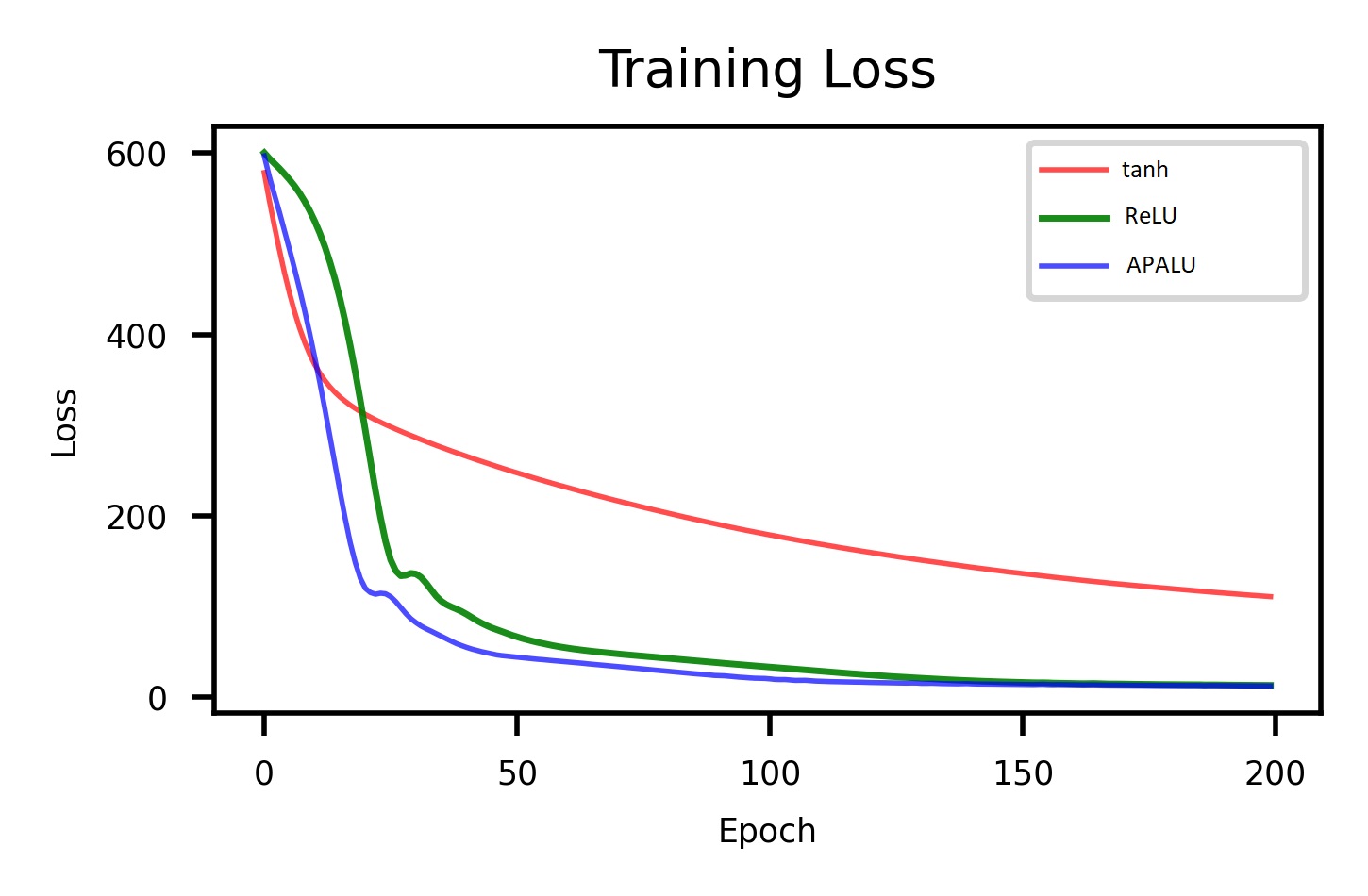}\\
    \includegraphics[width=.47\textwidth]{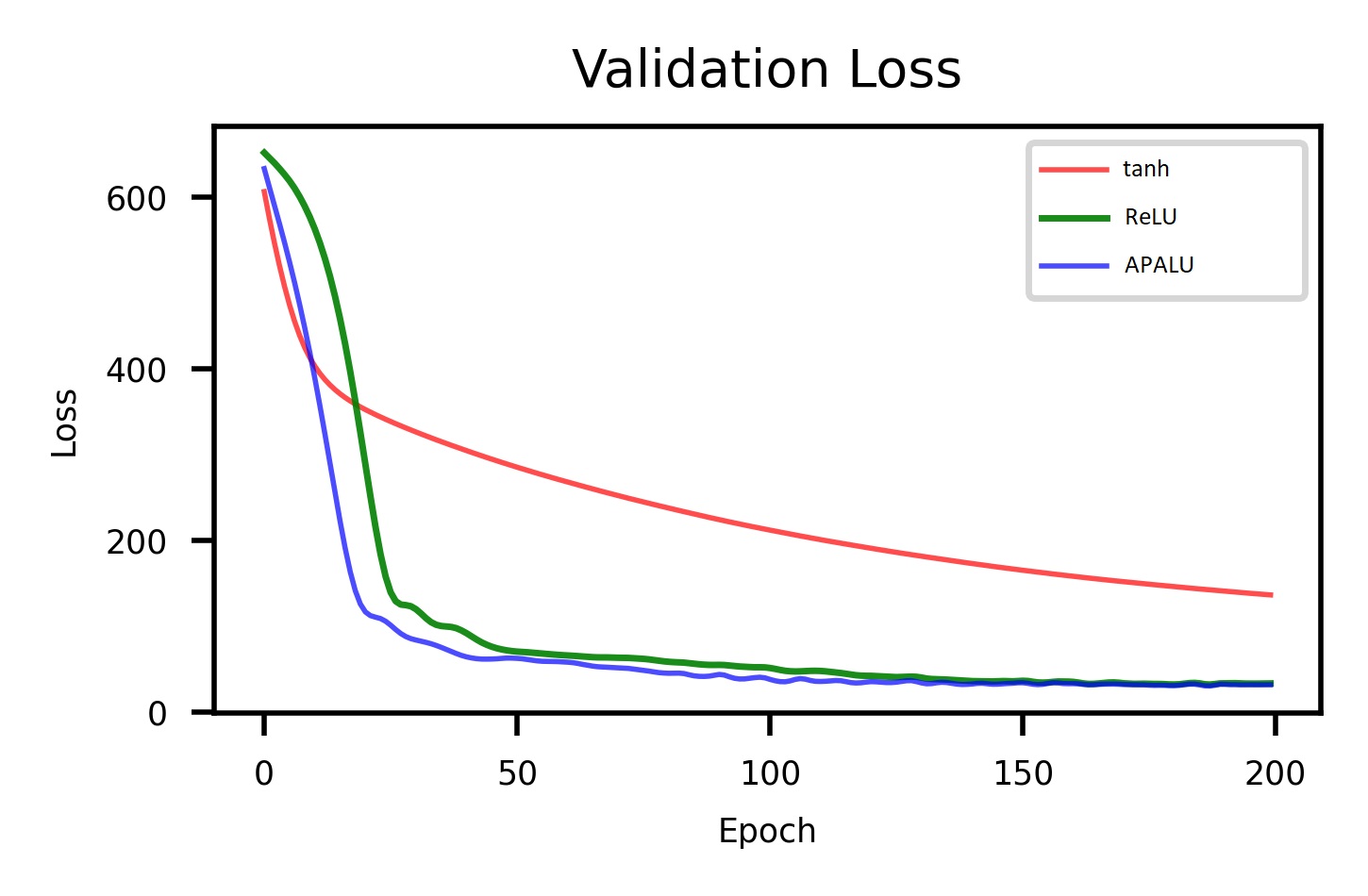}
    \caption{Training loss and test accuracy for MOPGRU model with different activation functions for sign language recognition task.}
 \end{figure} 

In summary, the APALU function, with carefully chosen parameters, has shown remarkable performance in various deep learning tasks compared to other activation functions. Its adaptive nature and the considered selection of $a$ and $b$ values have proven to be effective in various scenarios, underscoring its potential as a robust choice for different machine learning applications.

\section{Conclusion}
Activation functions play a crucial role in improving the performance deep learning models for better generalization. We propose a trainable activation function, APALU, to improve the performance of deep learning models on different tasks. Demonstrating remarkable adaptability, robustness, and enhanced performance, APALU outperformed classical activation functions across various tasks, including image classification to anomaly detection. The function's unique attributes, such as self-adjustability and differentiability, contribute to its effectiveness, while its sensitivity to hyperparameters and unexplored computational complexity pose avenues for future research. Further exploration into its generalization across diverse datasets and integration with various architectures will provide valuable insights and broaden its applicability. Overall, APALU offers a promising alternative to classical activation functions, with potential implications for a wide range of deep learning applications.

% \appendix

% \section*{Ethical Statement}

% There are no ethical issues.

% \section*{Acknowledgments}

% The preparation of these instructions and the \LaTeX{} and Bib\TeX{}
% files that implement them was supported by Schlumberger Palo Alto
% Research, AT\&T Bell Laboratories, and Morgan Kaufmann Publishers.
% Preparation of the Microsoft Word file was supported by IJCAI.  An
% early version of this document was created by Shirley Jowell and Peter
% F. Patel-Schneider.  It was subsequently modified by Jennifer
% Ballentine, Thomas Dean, Bernhard Nebel, Daniel Pagenstecher,
% Kurt Steinkraus, Toby Walsh, Carles Sierra, Marc Pujol-Gonzalez,
% Francisco Cruz-Mencia and Edith Elkind.

\bibliographystyle{named}
\bibliography{ijcai24}

\end{document}